\documentclass[runningheads]{llncs}
\usepackage[T1]{fontenc}
\usepackage{graphicx}
\usepackage{graphicx,verbatim}
\usepackage{amsmath}
\usepackage{amsfonts}
\usepackage{subcaption}
\usepackage{pgfplots}
\usepackage{url}
\usepackage{booktabs}
\begin{document}
\title{HU-based Foreground Masking for 3D Medical Masked Image Modeling}
%
%
\author{Jin Lee\inst{1}  \and
Vu Dang\inst{2} \and
Gwang-Hyun Yu\inst{1} \and 
Anh Le \inst{1} \and
Zahid Rahman\inst{1} \and
Jin-Ho Jang \inst{2} \and
Heonzoo Lee \inst{2} \and 
Kun-Yung Kim \inst{3} \and
Jin-Sul Kim \inst{1} \and
Jin-Young Kim \inst{1*}}
\authorrunning{J. Lee et al.}
%
\institute{Chonnam National University, South Korea \\
\email{\{ljin96,gwanghyunyu,anhle,zahid,jsworld,beyondi\}@jnu.ac.kr} \and
Research center, AISeed, Inc., South Korea \\
\email{\{dtvu,verycosy,heonzoo.lee\}@aiseed.kr} \and  
Seoul National University Bundang Hospital, South Korea \\
\email{kky2kkw@snubh.org} \\
* Corresponding author}
\maketitle              
\begin{abstract}
While Masked Image Modeling (MIM) has revolutionized fields of computer vision, its adoption in 3D medical image computing has been limited by the use of random masking, which overlooks the density of anatomical objects. To address this limitation, we enhance the pretext task with a simple yet effective masking strategy. Leveraging Hounsfield Unit (HU) measurements, we implement an HU-based Foreground Masking, which focuses on the intensity distribution of visceral organs and excludes non-tissue regions, such as air and fluid, that lack diagnostically meaningful features. Extensive experiments on five public 3D medical imaging datasets demonstrate that our masking consistently improves performance, both in quality of segmentation and Dice score (BTCV:~84.64\%, Flare22:~92.43\%, MM-WHS:~90.67\%, Amos22:~88.64\%, BraTS:~78.55\%). These results underscore the importance of domain-centric MIM and suggest a promising direction for representation learning in medical image segmentation. Implementation is available at \url{github.com/AISeedHub/SubFore/}.

\end{abstract}
\section{Introduction}\label{sec:introduction}

Masked Image Modeling (MIM) pretraining typically involves randomly selecting and masking regions of an image, followed by reconstructing the missing content. This strategy is well-suited to natural images\cite{xie2022simmim,he2022masked,bao2021beit}, where object brightness varies with lighting conditions, making the foreground independent of absolute pixel intensity. In contrast, computed tomography (CT) scans are acquired under fixed radiation exposure settings\cite{tang2022self,wu2024voco,dai2024sasamim}, and regions with low intensity—such as air or fluid—usually carry limited clinically relevant information.

\begin{figure}[!ht]
    \centering
    \begin{subfigure}[t]{0.3\textwidth}
        \includegraphics[width=\textwidth]{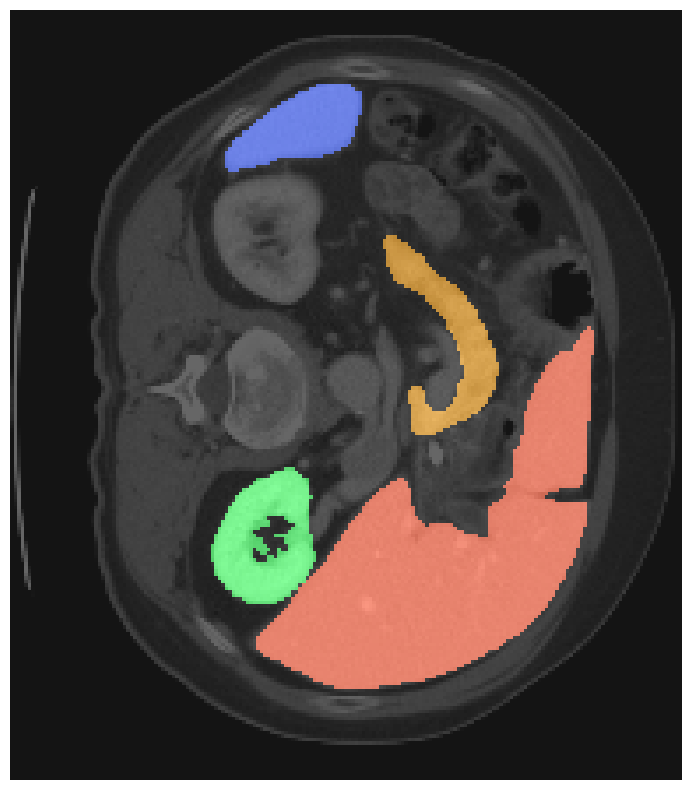} 
        \caption{Abdomen CT}\label{fig:left}
    \end{subfigure}
    \begin{subfigure}[t]{0.5\textwidth}
        \includegraphics[width=\textwidth]{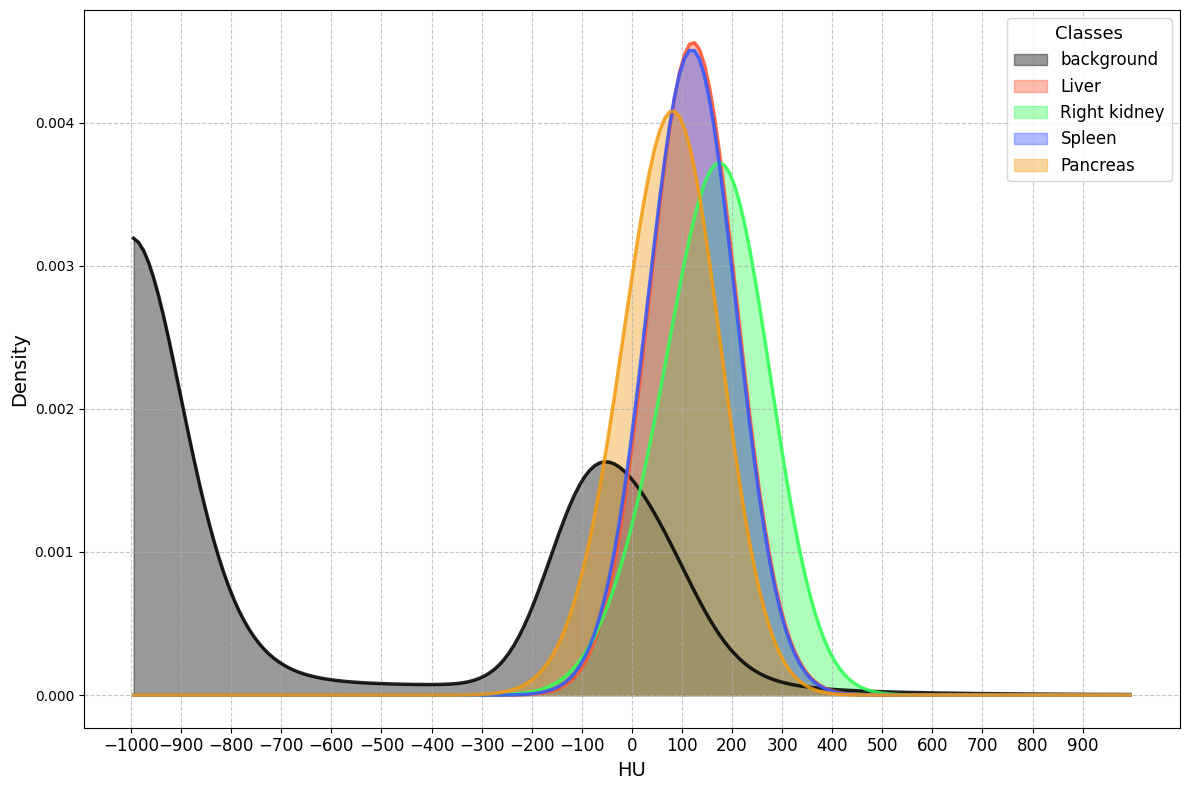} 
        \caption{HU value distribution}\label{fig:top_right}
    \end{subfigure} 
    \caption{Brightness distribution.~(a) An abdomen CT image with semantic segmentation labels.~(b) The distribution of HU values for the background and segmentation classes.}\label{fig:distribution}
\end{figure}

In CT imaging, voxel intensity values are measured in Hounsfield Units (HU), where -1000 HU corresponds to air, 0 HU to fluid spaces, and +1000 HU to dense tissues such as bone~\cite{ostmo2023view}. When converted into normalized pixel intensity for model input, these HU values reveal distinct intensity distributions. Our analysis of abdominal CT scans (Fig.\ref{fig:distribution}) shows that a large portion of the input with HU intensities less than 0, typically is background regions. These areas contribute little to the segmentation of soft or hard tissues.

Our core intuition is that applying random masking in the MIM paradigm to medical images is sub-optimal. Unlike natural scenes, medical scans often contain large homogeneous background regions lacking clinically meaningful structures. As such, their content can be more effectively modeled using structure-aware masking strategies. To address this, we propose HU-based Foreground Masking—a domain-specific approach that prioritizes organ-relevant regions in medical imaging. Specifically, we mask regions with intensity values within the interval $[0.1, 1]$, where informative content is densest, and discard those falling outside this range. This simple approach suprisingly outperforms serveral state-of-the-art baselines in the field, as discussed in section~\ref{sec:experiments}.

\section{Method}\label{sec:method}

\begin{figure}[!ht] 
    \centering
    \includegraphics[width=\linewidth,keepaspectratio]{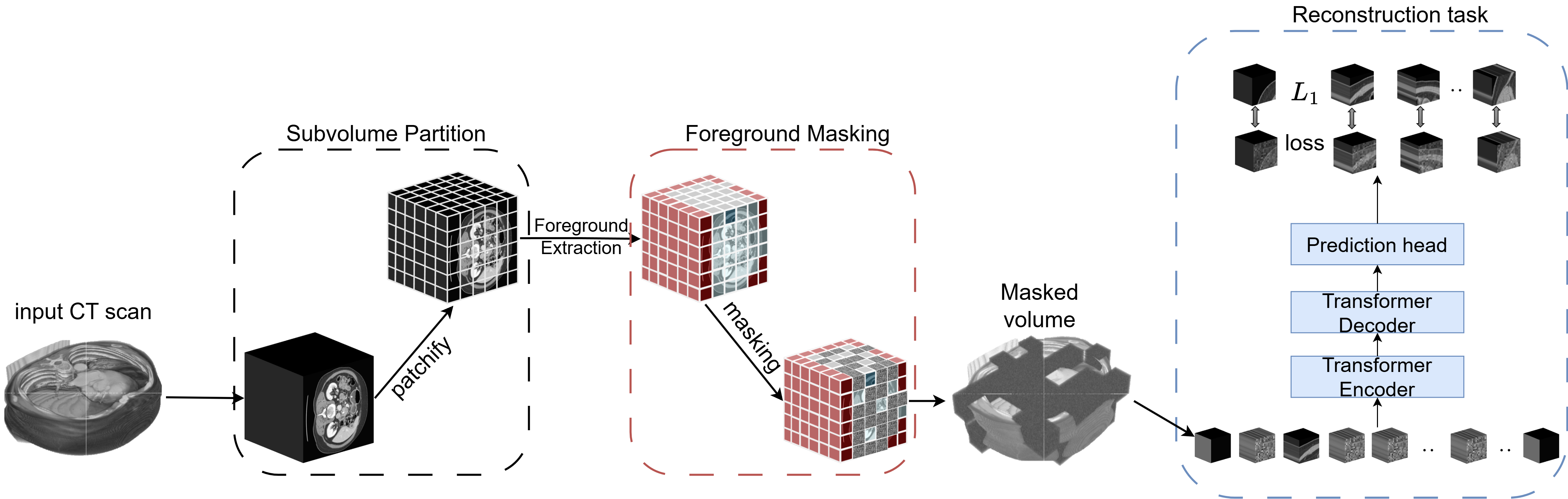}
    \caption{HU-based Foreground Masking framework. Given an input CT volume, our masking process consists of two steps: subvolume partition (in black) and foreground masking (in red). In the first step, the original volume is divided into multiple 3D subvolumes of size $16 \times 16 \times 16$. Then, the second step selectively masks subvolumes containing only the object regions. Finally, the pretext task (in blue) takes these subvolumes as inputs and learns to reconstruct the masked parts.}\label{fig:framework}
\end{figure}

We first empirically prove that object regions in 3D medical imaging are more informative than background regions (\ref{sec:analysis}). Then we introduce our effective masking method that involves two steps, subvolume partition, and foreground masking (\ref{sec:subvolume}) under the framework of Mask Image Modeling (Fig.\ref{fig:framework}).

\subsection{Object Regions in Medical Imaging}\label{sec:analysis}

We argue that, in medical imaging, foreground regions carry more diagnostically relevant information than background regions. For clarification, we compare object and background regions in terms of entropy, complexity, and mutual information (MI). Our analysis on the Flare22\cite{FLARE22-LDH2024} dataset show that object regions, exhibit significantly higher entropy, complexity, and MI than background regions, which often correspond to air or fluid with normalized HU values less than 0.1. These findings indicate that object regions dominate the informative content of medical images. So that excluding low-value background areas during pretraining not only preserves critical information but can also enhance model performance by focusing on anatomically meaningful structures.

\begin{figure}[!ht]
    \begin{subfigure}[t]{0.5\textwidth}
    \centering
        \begin{tikzpicture}
            \begin{axis}[
                ybar,
                bar width=10pt,
                symbolic x coords={MI, Entropy, Complexity},
                xtick=data,
                ylabel={Values (log scale)},
                ymin=0.01, ymax=8.0,
                legend style={at={(0.7, 0.9)},anchor=north,legend columns=1, font=\small},
                enlarge x limits=0.3,
                nodes near coords,
                every node near coord/.append style={font=\tiny, /pgf/number format/.cd, fixed, precision=2},
                width=\textwidth,
                height=5cm,
            ]
                \addplot coordinates {(MI, 5.64) (Entropy, 2.73) (Complexity, 0.35)};
                \addplot coordinates {(MI, 0.09) (Entropy, 0.02) (Complexity, 0.07)};
                \legend{Foreground, Background}
            \end{axis}
        
        \end{tikzpicture}
        \caption{}\label{fig:comparison_metrics}
    \end{subfigure}
    \begin{subfigure}[t]{0.5\textwidth}
        \includegraphics[width=\textwidth]{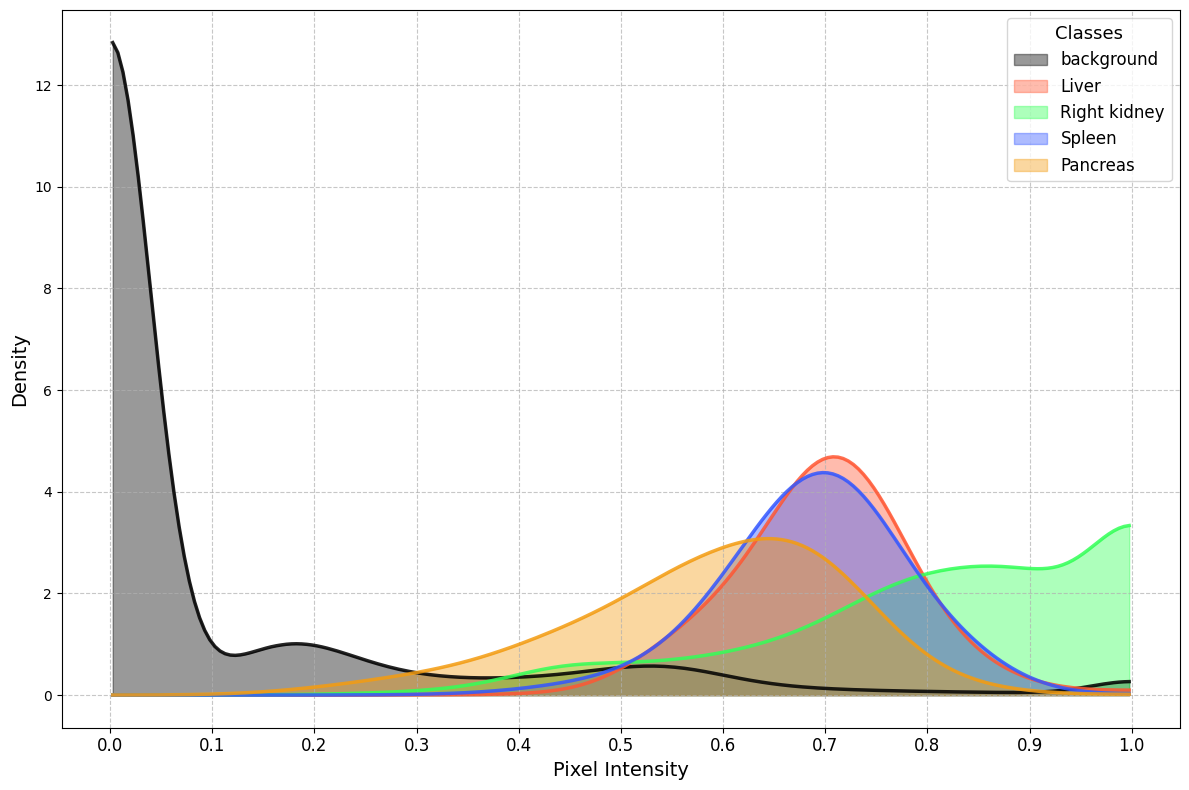}
        \caption{}\label{fig:Threshold}
    \end{subfigure}
    \caption{ROI analysis: (a) Comparison of mutual information, entropy, and complexity metrics for object and background regions; (b) The distribution of pixel intensities, where the brightness of anatomical objects lie in range $[0.1, 1]$.}
\end{figure}
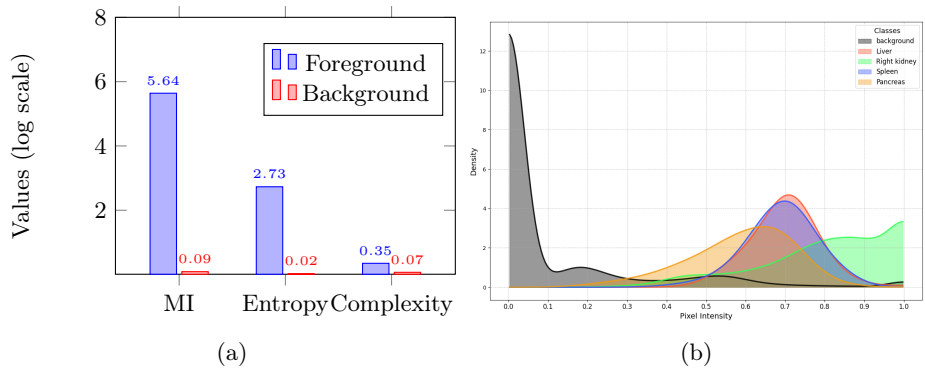

\subsection{HU-based Foreground Masking}\label{sec:subvolume}

\subsubsection{Subvolume Partition} In MIMs, visual signals are typically tokenized into patches. While methods like MAE~\cite{he2022masked} and SimMIM~\cite{xie2022simmim} use $16 \times 16$ image patches, volumetric models such as MAE3D~\cite{chen2023masked} extend them naively to 3D (e.g., $16 \times 16 \times 96$). Recent studies~\cite{wu2024voco,wu2024large,wald2025revisiting} show that dividing patches into smaller subvolumes along the depth improves performance. This approach enables better local context modeling and reduces embedding layer parameters—an advantage for high-dimensional medical volumes.

Formally, A 3D volume $\mathbf{X}$ can be divided into smaller non-overlapping subvolumes $\mathbf{X}_p$:

\begin{equation}
    \mathbf{X} = \bigcup_{p=1}^{P} \mathbf{X}_p
\end{equation}
where $P$ is the total number of subvolumes, and $X_p \in \mathbb{R}^{h \times w \times d}$ is the $p-th$ subvolume. Here, $h, w, d$ are the dimensions of the subvolume, satisfying $H =h n_h, W = w n_w, C =d n_d$ for $n_h, n_w, n_d \in \mathbb{Z}^{+}$ where $P = n_h  n_w  n_d$.

\subsubsection{Foreground Masking} Our foreground masking is defined over a subvolume $\mathcal{M_F}: \mathbf{X}_p \rightarrow \{0, 1\}$, where
\begin{equation}
    \mathcal{M_F}(\mathbf{X}_p) = \begin{cases}
        1,  & \text{if } \mathcal{H}(\mathbf{X}_p) \geq \lambda \\
        0,   & \text{otherwise}
    \end{cases}
\end{equation}
The masking function can either zero out or replace a subset of voxels based on a predefined strategy, such as random block masking\cite{chen2023masked,tang2022self} or semantic-driven masking\cite{valanarasu2023disruptive,wang2023hard,li2022semmae}. In this work, however, we adopt a relatively simple yet efficient approach that selectively retains only the foreground regions. Specifically, we define the characterized function $\mathcal{H}$ as the average of HU intensity belonging to the subvolume $\mathbf{X}_p$,
\begin{equation}
    \mathcal{H}(\mathbf{X}_p) = \frac{1}{h\times w \times d}\sum_{x=1}\sum_{y=1}\sum_{z=1}\mathbf{X}_p(x, y, z),
\end{equation}
The foreground subvolume is the one which characterized value higher than a predefined threshold $\lambda$. The masked input is then generated by applying $\mathcal{M_F}$ to all subvolumes of $\mathbf{X}$
\begin{equation}
    \mathbf{X}_{\mathcal{M}} = \mathcal{M}(\mathbf{X}) =  \bigcup_{p=1}^{P} \mathcal{M_F}(\mathbf{X}_p) \cdot \mathbf{X}_p 
\end{equation}

An analysis of 40 CT images identifies $\lambda = 0.1$ as the optimal threshold for separating anatomical structures from background regions. Fig.\ref{fig:Threshold} illustrates the smooth histogram distribution of background intensity compared to segmented organ labels. The HU intensity of the background is dense within low range, while the intensity of anatomical objects begins to appear around $0.1$ and is most prominent within the range $[0.4, 0.9]$.

\subsubsection{Pretext task} MIM's purpose is to learn a function $f_{\theta}$ which takes the masked volume $\mathbf{X}_{\mathcal{M}}$ as input:
\begin{equation}
    \hat{\mathbf{X}} = f_{\theta}(\mathbf{X}_{\mathcal{M}})
\end{equation}
and attempts to reconstruct the original volume $\mathbf{X}$. To train $f_{\theta}$, we use voxel-wise loss, $\mathcal{L}_{1}(\mathbf{X}, \hat{\mathbf{X}})$, to measure the discrepancy between the reconstructed $\hat{\mathbf{X}}$ and the original volume $\mathbf{X}$\cite{he2022masked,wang2023swinmm,kong2023understanding}. By restricting the loss to foreground masked indices, the model is encouraged to infer anatomical-driven representations.

\section{Experiment}\label{sec:experiments}
\subsection{Setting}\label{sec:setting}
Our settings are detailed as follows:

\textbf{Dataset:} To ensure comparability with previous studies\cite{wu2024voco,tang2022self,lee2024sdsl}, we use the three public datasets for pretraining: BTCV\cite{landman2015miccai}, TCIA Covid19\cite{clark2013cancer}, and LUNA\cite{setio2017validation}, comprising a total of approximately 1.6K CT scans. For downstream, we conduct experiments on five public datasets: BTCV (train: 24/valid: 6 CT scans), Flare22\cite{FLARE22-LDH2024} (train: 100/valid: 31 CT scans), MM-WHS\cite{zhuang2018multivariate} (train: 14/valid: 6 CT scans), Amos22\cite{ji2022amos} (train: 240/valid: 120 CT scans), and MSD BraTS (train: 388/valid: 96  MRI scans)\cite{simpson2019large}.

\textbf{Baseline:} We evaluate our masking with 2 backbones, UNETR\cite{hatamizadeh2022unetr} and SwinUNETR\cite{hatamizadeh2021swin}. For the UNETR backbone, we compare against the train-from-scratch model and two widely used MIM approaches, MAE\cite{he2022masked} and SimMIM\cite{xie2022simmim}. For the SwinUNETR backbone, we extend the comparison to include models pre-trained on medical datasets. With MIM-based baselines, we include pre-trained SwinUNETR\cite{tang2022self}, SwinMM\cite{wang2023swinmm}, DAE\cite{valanarasu2023disruptive}, and SDSL\cite{lee2024sdsl}. Additionally, we compare our method with VoCo\cite{wu2024voco}, a recent state-of-the-art volume-based contrastive pretraining.

\textbf{Configurations:} During pretraining, the UNETR model was optimized using AdamW with a learning rate of $10^{-4}$, while the SwinUNETR model used a learning rate of $8 \times 10^{-4}$. A weight decay of $0.05$ was applied to both models. No data augmentation was used during this stage. Training was conducted with a batch size of 4 and a fixed crop size of $(96, 96, 96)$ for 1600 epochs. For downstream segmentation tasks, the models were fine-tuned with a batch size of 2. All experiments, including pretraining and fine-tuning, were performed on a single NVIDIA H100 GPU.

\subsection{Downstream Evaluation on BTCV, Flare22 and MM-WHS}\label{sec:mainresult}

\begin{figure}[!ht]
    \centering

    \begin{subfigure}[t]{0.45\textwidth}
        \includegraphics[width=\textwidth]{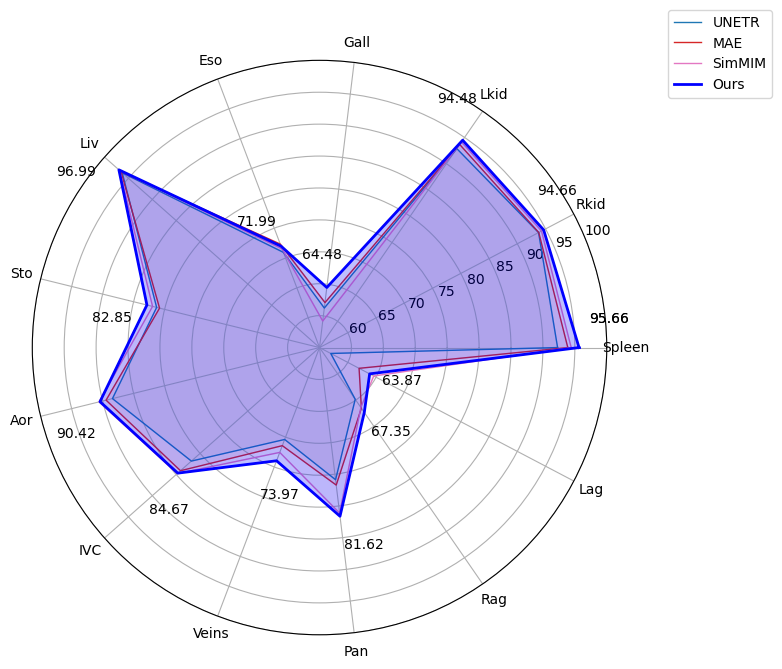}
        \caption{BTCV: UNETR backbone}\label{subfig:btcv_unetr}
    \end{subfigure}
    \hfill 
    \begin{subfigure}[t]{0.45\textwidth}
        \includegraphics[width=\textwidth]{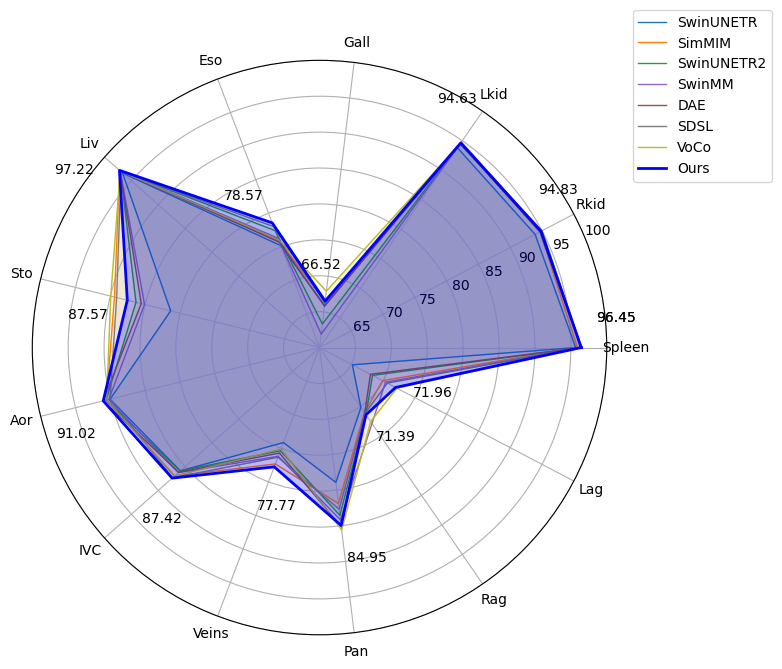}
        \caption{BTCV: SwinUNETR backbone}\label{subfig:btcv_swinunetr}
    \end{subfigure}


    \begin{subfigure}[t]{0.45\textwidth}
        \includegraphics[width=\textwidth]{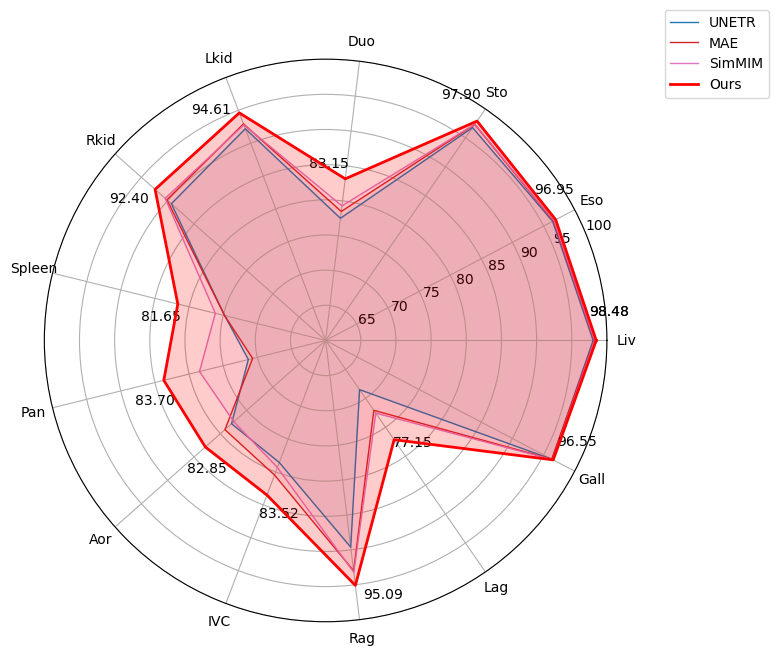}
        \caption{Flare22: UNETR backbone}\label{subfig:flare_unetr}
    \end{subfigure}
    \hfill
    \begin{subfigure}[t]{0.45\textwidth}
        \includegraphics[width=\textwidth]{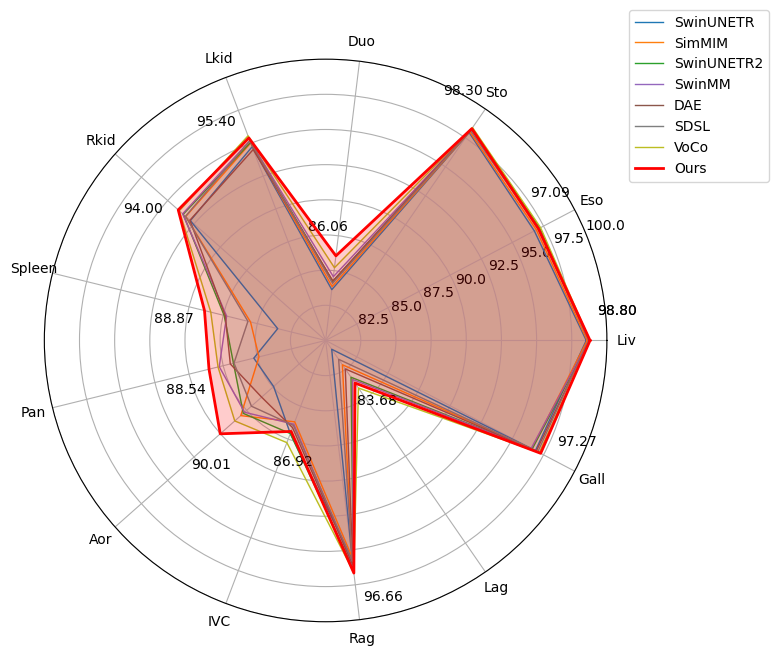}
        \caption{Flare22: SwinUNETR backbone}\label{subfig:flare_swinunetr}
    \end{subfigure}


    \begin{subfigure}[t]{0.45\textwidth}
        \includegraphics[width=\textwidth]{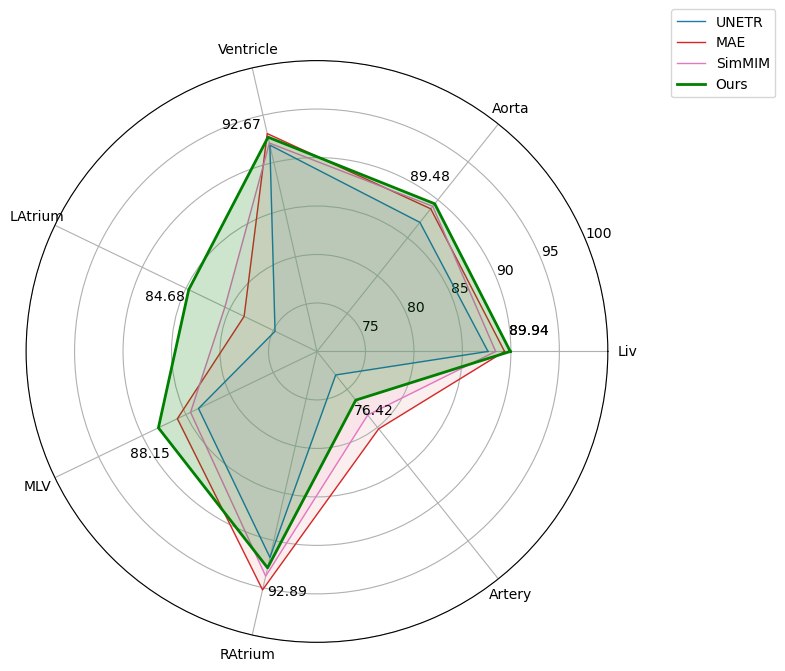}
        \caption{MM-WHS: UNETR backbone}\label{subfig:mmwhs_unetr}
    \end{subfigure}
    \hfill
    \begin{subfigure}[t]{0.45\textwidth}
        \includegraphics[width=\textwidth]{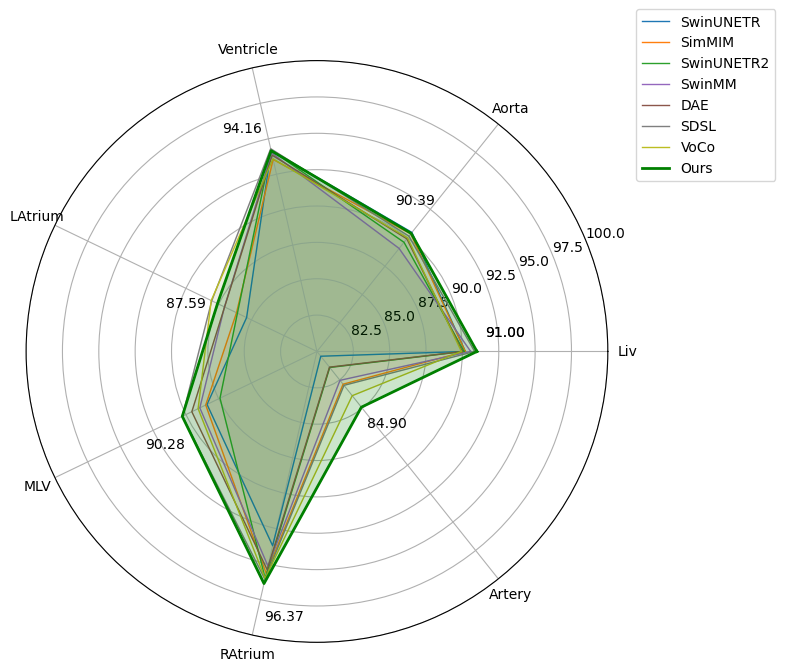}
        \caption{MM-WHS: SwinUNETR backbone}\label{subfig:mmwhs_swinunetr}
    \end{subfigure}

    \caption{Dice Score (\%) segmentation results across various datasets and backbones. (a)-(b) BTCV, (c)-(d): Flare22, (e)-(f): MM-WHS.}\label{fig:combined_segmentation_results}
\end{figure}

After pretraining with HU-based foreground masking strategy, we evaluate performance on downstream semantic segmentation tasks, where the model predicts voxel-wise labels for distinct anatomical structures. Breifly:

\begin{itemize}
    \item \textbf{BTCV:} The comparision of DSCs with other baselines is shown in Fig.\ref{subfig:btcv_unetr}-\ref{subfig:btcv_swinunetr}. Our approach achieves an average DSC of 81.7\% with the UNETR backbone, outperforming all baselines, and 84.64\% with the SwinUNETR backbone, achieving the best performance on 11 out of 12 tasks.
    \item \textbf{Flare22:} We achieve an average DSC of 89.54\% with the UNETR backbone and 92.43\% with the SwinUNETR backbone, outperforming all other methods. Notably, our approach yields substantial improvements in the segmentation of the pancreas, spleen, and aorta compared to baseline methods, as shown in Fig.\ref{subfig:flare_unetr}-\ref{subfig:flare_swinunetr}.
    \item \textbf{MM-WHS:} Using the UNETR backbone, our method achieves an average DSC of 87.75\%, ranking first on 5 out of 7 tasks. With the SwinUNETR backbone, it attains an average DSC of 90.67\%, outperforming all baseline methods across every task, as shown in Fig.\ref{subfig:mmwhs_unetr}-\ref{subfig:mmwhs_swinunetr}.
\end{itemize}

\begin{figure}[!ht]
    \centering
    \includegraphics[width=\linewidth,keepaspectratio]{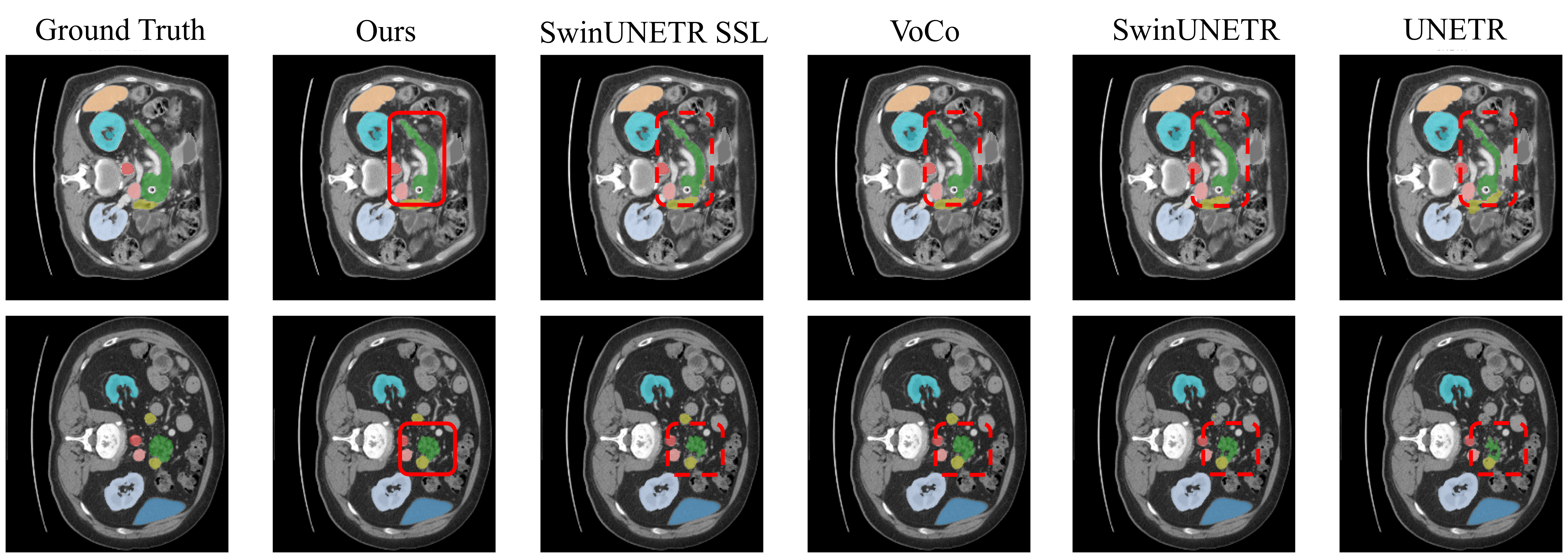}
    \caption{Segmentation quality comparison on Flare22 dataset. Our method shows better coverage of the pancreas marked in red box.}\label{fig:quality}
\end{figure}

Overall, our results reaffirm the advantages of pre-trained models, as our method surpasses the train-from-scratch baseline by a substantial margin for both UNETR ($2.91\%$) and SwinUNETR ($2.06\%$) backbones, averaged across three datasets. Using the UNETR backbone, our method achieves an average DSC improvement of $1.21\%$ over SimMIM\cite{xie2022simmim}, the second-best approach. Similarly, with the SwinUNETR backbone, our strategy is comparable to VoCo\cite{wu2024voco} ($0.32\%$), a state-of-the-art volumetric masking. We also show the improvement in segmentation quality (Fig.\ref{fig:quality}) where our approach delivers better coverage. Detailed Dice scores are available at:~\url{github.com/AISeedHub/SubFore/}.

\subsection{Ablation Study}\label{sec:ablation}

\begin{table}[h!]
\centering
\caption{Dice Score (\%) for Amos22 and BraTS datasets segmentation.}
\begin{tabular}{lccccc}
\toprule
\textbf{Methods} & \textbf{Amos22(Avg)} & \textbf{BraTS(TC)} & \textbf{BraTS(WT)} & \textbf{BraTS(ET)} & \textbf{BraTS(Avg)} \\
\midrule
\multicolumn{6}{c}{\textit{UNETR backbone}} \\
UNETR$\dagger$\cite{hatamizadeh2022unetr}       & 74.35       & 80.21     & 89.48     & 56.49     & 75.39      \\
MAE\cite{he2022masked}          & 76.94       & 79.89     & 89.51     & 56.69     & 75.36      \\
SimMIM\cite{xie2022simmim}      & 76.39       & 80.48     & 89.92     & 57.10     & 75.83      \\
Ours         & \textbf{77.34}       & \textbf{81.17}     & \textbf{90.19}     & \textbf{57.67}     & \textbf{76.34}      \\
\midrule
\multicolumn{6}{c}{\textit{Swin UNETR backbone}} \\
SwinUNETR$\dagger$\cite{hatamizadeh2021swin}    & 86.37       & 81.20     & 89.45     & 59.03     & 76.56      \\
SimMIM\cite{xie2022simmim}      & 87.27      & 82.72     & 90.63     & 60.10     & 77.82      \\
SwinUNETR\cite{tang2022self}  & 87.02       & 81.20     & 89.45     & 59.03     & 77.32      \\
SwinMM\cite{wang2023swinmm}       & 86.71       & 82.66     & 90.49     & 60.22     & 77.79      \\
DAE\cite{valanarasu2023disruptive}          & 87.04       & 82.47     & 90.31     & 59.71     & 77.50      \\
SDSL\cite{lee2024sdsl}        & 87.19       & 82.67     & 90.65     & 60.12     & 77.81      \\
VoCo\cite{wu2024voco}         & 88.49       & 82.66     & 90.78     & 60.46     & 77.97      \\
Ours         & \textbf{88.64}       & \textbf{83.46}     & \textbf{90.80}     & \textbf{61.39}     & \textbf{78.55}      \\
\bottomrule
\end{tabular}\label{tab:amos22_brats_segmentation_result}
\end{table}

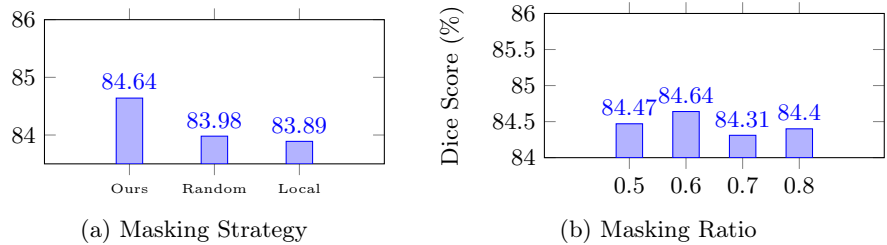
\begin{figure}[!ht]
    \begin{subfigure}[t]{0.5\textwidth}
        \centering
        \begin{tikzpicture}
            \begin{axis}[
                ybar,
                symbolic x coords={Ours, Random, Local},
                xtick=data,
                xticklabel style = {font=\tiny},
                ymin=83.5, ymax=86.0,
                bar width=10pt,
                enlarge x limits=0.5,
                nodes near coords,
                every node near coord/.append style={font=\small},
                width=\textwidth,
                height=3.5cm
            ]
                \addplot coordinates {
                    (Ours, 84.64)
                    (Random, 83.98)
                    (Local, 83.89)
                };
            \end{axis}
        \end{tikzpicture}
        \caption{Masking Strategy}\label{fig:chart2}
    \end{subfigure}
    \begin{subfigure}[t]{0.5\textwidth}
        \centering
        \begin{tikzpicture}
            \begin{axis}[
                ybar,
                symbolic x coords={0.5, 0.6, 0.7, 0.8},
                xtick=data,
                ylabel={Dice Score (\%)},
                ymin=84.0, ymax=86.0,
                bar width=10pt,
                enlarge x limits=0.5,
                nodes near coords,
                every node near coord/.append style={font=\small},
                width=\textwidth,
                height=3.5cm
            ]
                \addplot coordinates {(0.5, 84.47) (0.6, 84.64) (0.7, 84.31) (0.8, 84.40)};
            \end{axis}
        \end{tikzpicture}
        \caption{Masking Ratio}\label{fig:chart1}
    \end{subfigure}
    
    \caption{Ablation Study.}\label{fig:ablation}
\end{figure}

\subsubsection{Adaptability} We further evaluated our method on a larger labeled dataset, AMOS, which contains 360 annotated CT scans. Moreover, despite being pretrained exclusively on CT data, our masking strategy generalizes well to MRI images on the BraTS dataset. Our method proves its adaptability when achieves higher DSCs across all segmentation tasks (Table~\ref{tab:amos22_brats_segmentation_result}).

\subsubsection{Robustness} Fig.\ref{fig:ablation} presents ablation study on two pretraining factors, masking strategy and masking ratio. Each experiment was performed on the BTCV dataset and the results were reported in the average Dice score. The best-performing model was trained with our HU-based foreground masking strategy, using a masking ratio of 0.6. Compared to random\cite{chen2023masked}, and local\cite{valanarasu2023disruptive} masking, our method yielded the significant improvement (0.75\% in DSC), profound the importance of semantic-driven approaches in designing masking methods.


\section{Conclusions}

Anatomical intensity in medical images offer valuable cues that can be leveraged for pretraining phase in Masked Image Modeling. In this study, we employ HU-based foreground masking, an efficient masking way that improves performance across five public medical imaging datasets for multi-organ and brain tumor segmentation. However, we acknowledge a limitation in our foreground extraction itself when using a single global threshold may not be optimal across subvolumes and datasets. This limitation suggests an avenue for future research to explore adaptive thresholding techniques.

\begin{credits}
\subsubsection{\ackname} This work was partly supported by the Innovative Human Resource Development for Local Intellectualization program through the Institute of Information \& Communications Technology Planning \& Evaluation (IITP) grant funded by the Korean government (MSIT) (IITP-2024-RS-2022-00156287, 50) and this work was partly supported by an Institute of Information \& Communications Technology Planning \& Evaluation (IITP) grant funded by the Korean government (MSIT) (RS-2021-II212068, Artificial Intelligence Innovation Hub) and this work was supported by Artificial intelligence industrial convergence cluster development project funded by the Ministry of Science and ICT(MSIT, Korea)\&Gwangju Metropolitan City.

\subsubsection{\discintname}
The authors have no competing interests to declare that are
relevant to the content of this article.

\end{credits}
%
%
%
\bibliographystyle{splncs04}
\bibliography{mybibliography}
%




\end{document}